# *It's Too Many Options*: Pitfalls of Multiple-Choice Questions in Generative AI and Medical Education

Shrutika Singh[1], Anton Alyakin[1,2,3], Daniel Alexander Alber[1,4], Jaden Stryker[1], Ai Phuong S Tong[5], Karl Sangwon[1,4], Nicolas Goff[1], Mathew de la Paz[3], Miguel Hernandez-Rovira[2,3], Ki Yun Park[2,3], Eric Claude Leuthardt[2], Eric Karl Oermann [1,6,7,8]

[1] Department of Neurosurgery, NYU Langone Health, New York, NY, USA.
[2] Department of Neurosurgery, Washington University in Saint Louis, Saint Louis, Missouri, USA
[3] Washington University in Saint Louis School of Medicine, Saint Louis, Missouri, USA
[4] New York University Grossman School of Medicine, New York, NY, USA
[5] University of Washington School of Medicine, Seattle, Washington, USA.
[6] Center for Data Science, New York University, New York, NY, USA.
[7] Department of Radiology, NYU Langone Health, New York, NY, USA.
[8] Neuroscience Institute, NYU Langone Health, New York, NY, USA

**Corresponding author during review**
Shrutika Singh
NYU Langone Department of Neurosurgery
550 First Ave
New York, NY 10016
ss17775@nyu.edu

**Corresponding author post-review**
Eric K. Oermann
NYU Langone Department of Neurosurgery
550 First Ave
New York, NY 10016
Eric.Oermann@nyulangone.org

**Abstract**

The performance of Large Language Models (LLMs) on multiple-choice question (MCQ) benchmarks is frequently cited as proof of their medical capabilities. We hypothesized that LLM performance on medical MCQs may in part be illusory and driven by factors beyond medical content knowledge and reasoning capabilities. To assess this, we created a novel benchmark of free-response questions with paired MCQs (FreeMedQA). Using this benchmark, we evaluated three state-of-the-art LLMs (GPT-4o, GPT-3.5, and LLama-3-70B-instruct) and found an average absolute deterioration of 39.43% in performance on free-response questions relative to multiple-choice ($p = 1.3 * 10^{-5}$) which was greater than the human performance decline of 22.29%. To isolate the role of the MCQ format on performance, we performed a masking study, iteratively masking out parts of the question stem. At 100% masking, the average LLM multiple-choice performance was 6.70% greater than random chance ($p = 0.002$) with one LLM (GPT-4o) obtaining an accuracy of 37.34%. Notably, for all LLMs the free-response performance was near zero. Our results highlight the shortcomings in medical MCQ benchmarks for overestimating the capabilities of LLMs in medicine, and, broadly, the potential for improving both human and machine assessments using LLM-evaluated free-response questions.

**Introduction**
In recent years, foundation models have become more prevalent in medical research[1–4]. These models have purportedly demonstrated high performance across a variety of fields in medicine, and easily passed formal assessments of medical knowledge such as the United States Medical Licensing Exam[2–4]. One of the more prominent benchmarks used to report the performances of LLMs is the MultiMedQA which encompasses questions from many fields and stages of medical training[2,5–7]. Notably, this benchmark and others are composed of multiple-choice questions, which may present limitations for the accurate assessment of LLMs[8–10]. While recent works such as CRAFT-MD have focused on converting these multiple-choice questions into more real-world assessments involving multi-turn conversations, there are still no rigorous evaluations of the quality of these multiple-choice benchmarks themselves[11].

We hypothesized that existing multiple-choice question (MCQ) benchmarks are poor metrics for assessing the medical knowledge and capabilities of LLMs. To test this, we developed a benchmark of paired free-response and multiple-choice questions and developed a technique for automatically assessing free-response answers. We then compared the performance of GPT-4o [12], GPT-3.5[13], and Llama-3-70B[14] to answer questions when presented in both multiple-choice and free-response formats. We further studied the performance of these LLMs when the question stems were progressively masked in both free-response and multiple-choice formats. We hypothesized that multiple-choice performance should approach random chance at 25% as information is increasingly lost to masking. Lastly, we conducted human evaluations with medical students to establish human baselines and provide context for LLM results.

**Results**
*FreeMedQA creation*
Starting with 14,965 candidate questions from the MultiMedQA and using an LLM-based pipeline (see Methods, **Extended Data Fig. 1**), we created 10,278 questions with paired free-response and MCQ versions (FreeMedQA). We also built an evaluative method using GPT-4o as a judge to score free-response answers based on MCQ answers (**Extended Data Fig. 2**).

*Evaluation of LLMs' performance in free-response compared to multiple-choice*
Using this novel benchmark, we found that GPT-4o, GPT-3.5, and Llama-3-70B-Chat report significant drops in performance when evaluated using a free-response question format as opposed to MCQ. On average, the models attained a 39.43% (combined p = 1.3 * $10^{-5}$) absolute drop in performance from multiple-choice to free-response answering capabilities Llama 3 acquired the greatest absolute drop of 46.59% (relative drop of 59.08%; p = 0.006), followed by GPT-4o reporting an absolute 37.50% drop (relative drop of 43.23%;p = 0.004), and GPT-3.5 reporting the lowest drop at 34.20% (relative drop of 56.51%; p = 0.004) (**Fig. 1**).

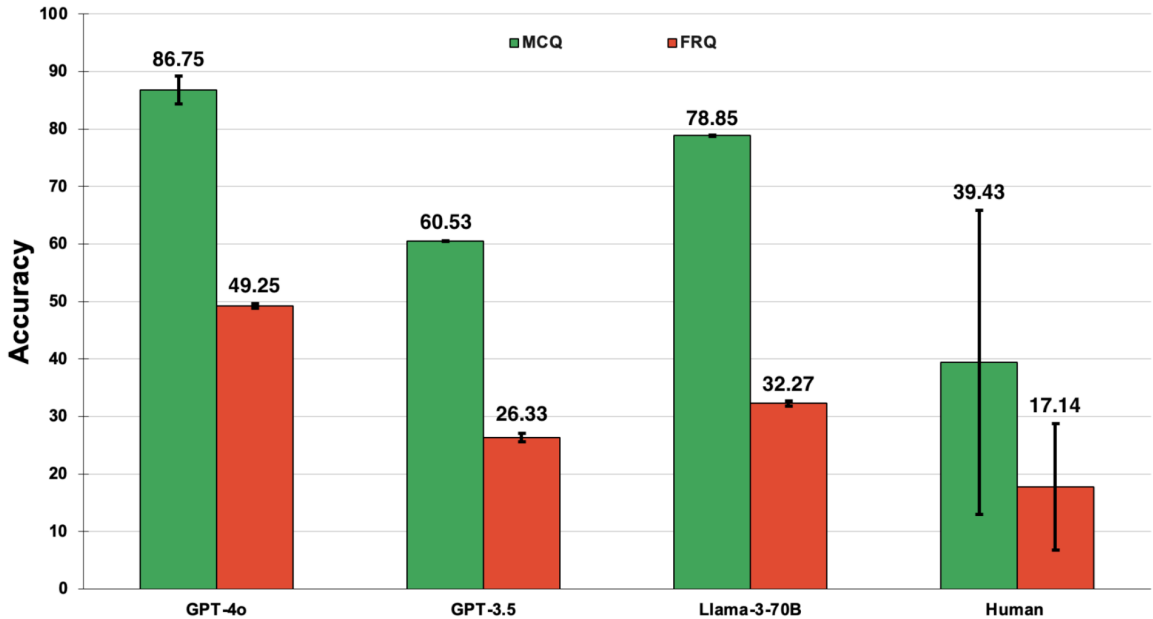

**Fig. 1. LLMs performance on FreeMedQA.** Performance of gpt-4o-2024-08-06, gpt-3.5-turbo-0125, llama3-70B-chat and Medical Students on FreeMedQA. All models displayed depreciated performances in the free-response category when compared to the multiple-choice with a 39.43% average drop in performance. Notably this drop in performance is similar to that of humans who had a 22.29% decline in performance with the transition from multiple-choice to free-response. Error bars represent the standard deviation obtained from repeating the experiment.

*Evaluation of medical students' performance in free-response compared to multiple-choice*
To contextualize these findings we also assessed medical trainees using FreeMedQA. We found that senior medical students experienced a 22.29% decrease in performance when transitioning from multiple-choice to free-response questions ($p = 0.008$), with scores dropping from 39.43% on multiple-choice to 17.79% on free-response questions (**Fig. 1**).

*Evaluation of LLMs' performance with masked inputs*
To investigate this relative performance of LLMs further, we performed a masking study where we progressively masked out the question stems of FreeMedQA questions. For the multiple-choice component, the answer options were presented void of any masking. All models deteriorated in performance in both multiple-choice and free-response categories as increasing portions of the input were masked. A notable discrepancy is at 100% masking where the multiple-choice performance is on average 6.70% greater than that of a random chance of 25%. GPT-4o had the greatest deviation from random chance with an accuracy of 37.34%, 12.34% higher than random chance, despite complete masking of the inputs ($p = 0.031$). Comparatively, across all models, the free-response performance declines to 0.15%, a value slightly greater than zero due to noise introduced by the evaluating model: GPT-4o (**Fig. 2**).

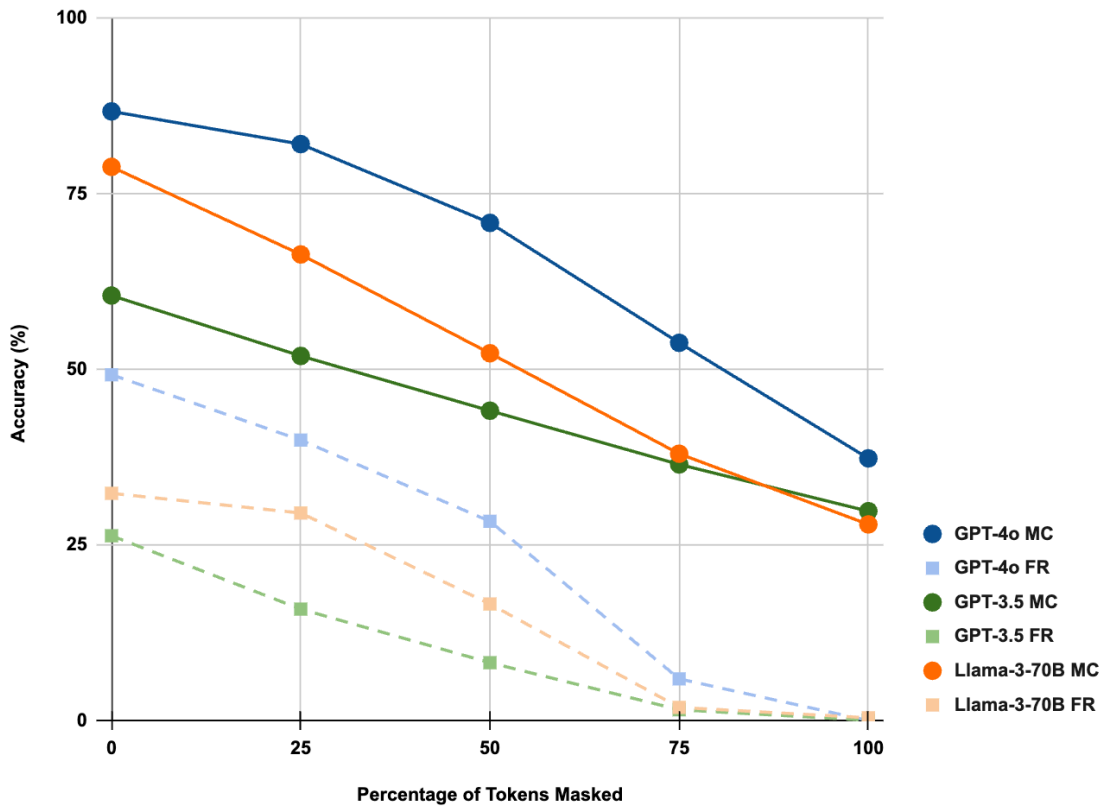

**Fig. 2. Performance of LLMs with Masking.** A graphical presentation of the performance of the studied models, when the prompt is masked in 25% increments, in free-response and multiple-choice formats. At 100% masking, all models report an accuracy greater than the possibility of random chance i.e. 25% in the multiple-choice category whereas the accuracy of the free-response category plummets to zero, implying an innate process of pattern recognition being used by the LLMs.

**Discussion:**
We present a straightforward revision of existing multiple-choice benchmarks for medical LLMs – converting them to free-response questions – that may aid researchers and clinicians in elucidating LLMs' strengths and weaknesses. We find that medical LLMs have learned processes to determine the answer in a multiple-choice setting that is independent of their ability to answer the question being asked, as their performance declines significantly but remains above chance when input information is masked up to the entire input. Therefore, medical LLM performance on benchmarks composed of multiple-choice questions does not reflect their genuine understanding of medical concepts in a more general setting, and reformulating LLM assessments as free-response questions or multi-turn dialogues[11] seems prudent.

A degradation in performance in the free-response format compared to the multiple-choice structure across all models was also observed. We attribute this to learned mechanisms utilized by LLMs to recognize the answer reliant solely on the answer options. Interestingly, this performance decrease is similar to that of humans, suggesting that some of this gap is due to a more general test-taking strategy that is leveraged by both medical LLMs and medical trainees. In fact, a parallel could be drawn from the different levels of learning witnessed in humans via Bloom's Taxonomy [15] to similar mechanisms in LLM learning. Mere recognition of the correct choice becomes a more feasible task than recalling all associated information.

Other recent works have echoed this concern over the use of multiple-choice questions for assessing medical LLMs[16]. The recently released CRAFT-MD benchmark adapts medical multiple-choice questions to multi-turn dialogues, which poses an interesting alternative to free-response questions[11]. Also, calls to consider evaluating medical LLMs using processes similar to medical trainees seem to be increasingly justified in light of such results[17].

This study is not without limitations. To maintain the integrity of the study, we removed 31.32% of the questions in the MultiMedQA that required knowledge of the answer options to identify the correct answer. This decreased the size of our derived FreeMedQA benchmark and also raised more general concerns over test question quality. Our study was conducted only on a subset of one popular medical benchmark, but there are other popular medical benchmarks used to report LLM performance.[18] This study was restricted to the English language, but medical benchmarks in other languages also exist[19]. Lastly, our FreeMedEval approach utilizes GPT-4o, which while efficient, is stochastic and not void of error although GPT-4 is commonly utilized in this manner in other studies[11]. The evaluation of natural language generation is an open problem in natural language processing which we leave for future work. For future avenues of investigation, this study could be conducted on a dataset specific to a specialty in medicine[20,21]. The study could be scaled to additional modalities or scaled to include multiple popular datasets[22–24]. Lastly, further efforts to improve the evaluation of natural language generation and free-response answers are sorely needed.

**Conclusion**
Our study and other recent studies show that the evaluation of medical LLMs and medical AI in general is clearly an open problem. The push to develop and introduce AI models into clinical care needs to be met with equally strong and creative approaches to appropriately evaluate these models to ensure their safety and reliability.

**References**

1. Bommasani, R. *et al.* On the Opportunities and Risks of Foundation Models. *arXiv [cs.LG]* (2021).

2. Singhal, K. *et al.* Large language models encode clinical knowledge. *Nature* **620**, 172–180 (2023).

3. Nori, H. *et al.* Can generalist foundation models outcompete special-purpose tuning? Case study in medicine. *arXiv [cs.CL]* (2023).

4. Nori, H., King, N., McKinney, S. M., Carignan, D. & Horvitz, E. Capabilities of GPT-4 on Medical Challenge Problems. *arXiv [cs.CL]* (2023).

5. Blanco, J., Lambert, C. & Thompson, O. GPT-Neo with LoRA for better medical knowledge performance on MultiMedQA dataset. (2024) doi:10.31219/osf.io/njupy.

6. Bolton, E. *et al.* Assessing The Potential Of Mid-Sized Language Models For Clinical QA. *arXiv [cs.CL]* (2024).

7. Hamzah, F. & Sulaiman, N. Optimizing llama 7B for medical question answering: A study on fine-tuning strategies and performance on the MultiMedQA dataset. https://osf.io/g5aes/download.

8. Li, W. *et al.* Can multiple-choice questions really be useful in detecting the abilities of LLMs? *arXiv [cs.CL]* (2024).

9. Balepur, N. & Rudinger, R. Is Your Large Language Model Knowledgeable or a Choices-Only Cheater? *arXiv [cs.CL]* (2024).

10. Schubert, M. C., Wick, W. & Venkataramani, V. Performance of Large Language Models on a Neurology Board–Style Examination. *JAMA Netw Open* **6**, e2346721–e2346721 (2023).

11. Johri, S. *et al.* An evaluation framework for clinical use of large language models in patient interaction tasks. *Nat. Med.* 1–10 (2025).

12. OpenAI *et al.* GPT-4 Technical Report. *arXiv [cs.CL]* (2023).

13. Brown, T. *et al.* Language models are few-shot learners. *Adv. Neural Inf. Process. Syst.* **33**,

1877–1901 (2020).

14. Grattafiori, A. *et al.* The Llama 3 herd of models. *arXiv [cs.AI]* (2024).

15. Bilon, E. *Using Bloom's Taxonomy to Write Effective Learning Objectives: The Abcds of Writing Learning Objectives: A Basic Guide*. (Independently Published, 2019).

16. Griot, M., Vanderdonckt, J., Yuksel, D. & Hemptinne, C. Multiple Choice Questions and Large Languages Models: A Case Study with Fictional Medical Data. *arXiv [cs.CL]* (2024).

17. Rajpurkar, P. & Topol, E. J. A clinical certification pathway for generalist medical AI systems. *Lancet* **405**, 20 (2025).

18. Xu, J. *et al.* Data Set and Benchmark (MedGPTEval) to Evaluate Responses From Large Language Models in Medicine: Evaluation Development and Validation. *JMIR Med Inform* **12**, e57674 (2024).

19. Cai, Y. *et al.* MedBench: A Large-Scale Chinese Benchmark for Evaluating Medical Large Language Models. *AAAI* **38**, 17709–17717 (2024).

20. Longwell, J. B. *et al.* Performance of Large Language Models on Medical Oncology Examination Questions. *JAMA Netw Open* **7**, e2417641 (2024).

21. Pellegrini, C., Keicher, M., Özsoy, E. & Navab, N. Rad-ReStruct: A Novel VQA Benchmark and Method for Structured Radiology Reporting. in *Medical Image Computing and Computer Assisted Intervention – MICCAI 2023* 409–419 (Springer Nature Switzerland, 2023).

22. Adams, L. *et al.* LongHealth: A Question Answering Benchmark with Long Clinical Documents. *arXiv [cs.CL]* (2024).

23. Dada, A. *et al.* CLUE: A Clinical Language Understanding Evaluation for LLMs. *arXiv [cs.CL]* (2024).

24. Chen, Q. & Deng, C. Bioinfo-Bench: A Simple Benchmark Framework for LLM Bioinformatics Skills Evaluation. *bioRxiv* 2023.10.18.563023 (2023) doi:10.1101/2023.10.18.563023.

25. MultiMedQA - a openlifescienceai Collection. https://huggingface.co/collections/openlifescienceai/multimedqa-66098a5b280539974cefe485.

## Methods

*FreeMedQA Filtering*

We utilized the MultiMedQA dataset acquired from the Hugging Face Hub[25]. For each question, we prompted GPT-4o for whether it could be answered without multiple-choice options included. We employed few-shot prompting for this task with ten manually curated examples. This process yielded a subset of 10,278 MultiMedQA questions which were deemed appropriate to be converted to a free response format. The multiple-choice versions of the dataset composed the MC portion of our new FreeMedQA dataset. (**Extended Data Fig. 1.**)

*Filtering Quality Control*

A manual review was conducted to evaluate GPT-4o's ability to categorize questions as answerable or not without answer choices, a manual review was conducted. 100 random questions were sampled from the MultiMedQA dataset, and judged to require answer options or not by a senior medical student who was blinded to the GPT-4o's decision. (**Extended Data Fig. 2.**) The result of the manual review led to the conclusion that GPT-4o tends to be conservative, but suitable for the task.

*Free-Response Adaptation*

We employed regular expressions (RegEx) string matching to identify and replace phrases that are specific to multiple-choice questions. For example, "which of the following" was replaced with "what" to align with the free-response structure. The resulting questions with the corresponding correct answers composed the free response (FR) portion of our FreeMedQA dataset.

*Performance Assessment*

We performed a comparative study of three different industry-grade LLMs - GPT-4o, GPT-3.5, and Llama-3-70B. We first prompted each model to answer the multiple-choice version of the question in FreeMedQA and evaluated the result using string matching. We used a maximum context length of 1024, and a temperature of 0.0, and repeated each experiment five times to obtain statistical bounds. We then presented each model with a free-response version of each question. We used GPT-4o to evaluate the correctness of the answer by presenting it with the correct answer choice and the candidate's answer choice and prompting it to evaluate if two answers are similar (one is contained in the other) (**Extended Data Fig. 3**). Notably, it was blinded to the question being asked. We similarly repeated the experiment five times.

We obtained error bars by computing standard deviations over the reruns of our experiments. (**Fig. 1**) We established statistical significance by using the Mann-Whitney *U* test between the five of each repeat. We aggregated the *p* values using Fisher's method.

*Masked Study*

We systematically masked portions of the questions in increasing increments of 25%, hiding progressively larger sections of the question stemming from the model. For the multiple-choice format, the answer choices remained fully visible at all masking levels.

We obtained standard deviations used for the error bars by repeating each experiment a total of five times. To establish statistical significance in the multiple choice with 100% masking study, we used the Wilcoxon signed-rank test with a null hypothesis of random chance probability of 25% and an alternative that models perform better than chance.

*Medical Student Knowledge Evaluation*
From each dataset, a random selection of 200 questions was made, resulting in a total of 350 unique, non-overlapping questions. These questions were distributed across eight Google Forms, each containing 50 questions, 25 multiple-choice and 25 free-response. The forms featured a randomized arrangement of multiple-choice and free-response questions. The multiple-choice questions were evaluated by comparison to the correct answer stored in the FreeMedQA by GPT-4o. The free-response questions were assessed via synonymity assessment by GPT-4o. Five senior medical students participated in completing these forms. We performed a one-sided paired Wilcoxon signed-rank test on multiple choice vs free response averages achieved on every form by the med students with an alternative of better performance on the multiple-choice forms.

**Extended Data**

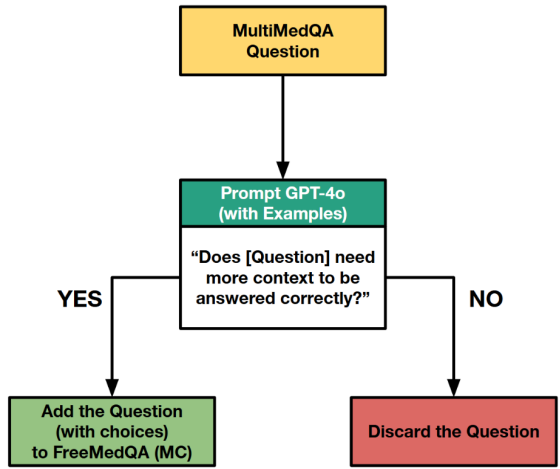

**Extended Data Figure 1. Dataset Formation.** We prompted GPT-4o to determine if questions were capable of being answered without multiple-choice options. This process was conducted on all the 14,965 questions in the MultiMedQA. The output of which was a 10,278 subsection, the FreeMedQA.

**Human Evaluation**

|  | Include | Exclude |
|---|---|---|
| **GPT-4o Suggestion: Include** | 67 | 18 |
| **GPT-4o Suggestion: Exclude** | 10 | 5 |

**Extended Data Figure 2. Verification of GPT-4o's Assessment.** A manual review of GPT-4o's ability to assess the need for additional context for a question was conducted.

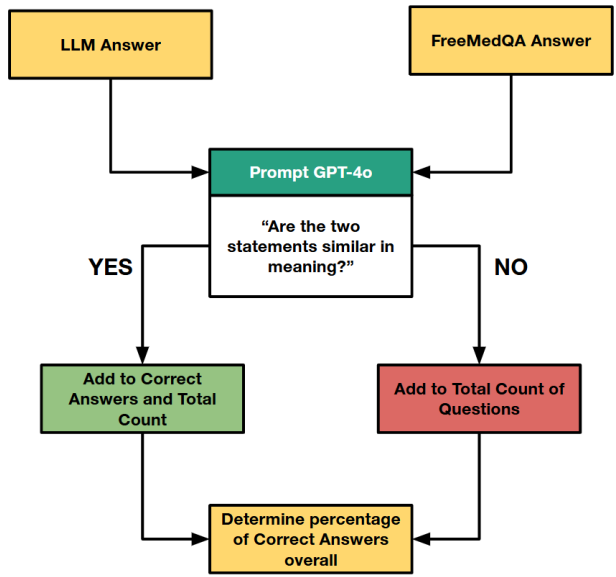

**Extended Data Figure 3. Free-response Evaluation Process.** We prompted GPT-4o to determine the similarity between the LLM Answer and the correct answer as per the FreeMedQA to evaluate the free-response performance.

| Large Language Models | multiple-choice | free-response |
|---|---|---|
| gpt-4o-2024-08-06 | 86.75 ± 2.41 | 49.25 ± 0.40 |
| gpt-3.5-turbo-01235 | 60.53 ± 0.05 | 26.33 ± 0.75 |
| llama-3-70B-chat | 78.85 ± 0.11 | 32.27 ± 0.43 |

**Extended Data Table 1. Results of running LLMs on FreeMedQA Dataset of 10,278 Questions.** We assessed the performance of the respective large language models (LLMs) on the FreeMedQA dataset, comprising 10,278 questions. The evaluation revealed a significant decline in accuracy from multiple-choice to free-response format, highlighting the inability of multiple-choice questions to appropriately assess understanding.

| Large Language Models | 0% Masking | | 25% Masking | | 50% Masking | | 75% Masking | | 100% Masking | |
|---|---|---|---|---|---|---|---|---|---|---|
| | multiple-choice | free-response | multiple-choice | free-response | multiple-choice | free-response | multiple-choice | free-response | multiple-choice | free-response |
| gpt-4o-2024-08-06 | 86.75 ± 2.41 | 49.25 ± 0.40 | 82.09 ± 0.09 | 39.95 ± 0.85 | 70.88 ± 0.05 | 28.39 ± 0.22 | 53.80 ± 0.07 | 5.95 ± 0.08 | 37.34 ± 0.20 | 0.01 ± 0.00 |
| gpt-3.5-turbo-01235 | 60.53 ± 0.05 | 26.33 ± 0.75 | 51.93 ± 0.10 | 15.88 ± 0.19 | 44.13 ± 0.05 | 8.25 ± 0.19 | 36.48 ± 0.07 | 1.55 ± 0.06 | 29.82 ± 0.10 | 0.04 ± 0.01 |
| llama-3-70B-chat | 78.85 ± 0.11 | 32.27 ± 0.43 | 66.38 ± 0.33 | 29.56 ± 0.44 | 52.31 ± 0.08 | 16.61 ± 0.20 | 38.00 ± 0.22 | 1.88 ± 0.14 | 27.94 ± 0.43 | 0.39 ± 0.05 |

**Extended Data Table 2. LLM results of masking increasing percentages of the prompt on the FreeMedQA Dataset.** We evaluated the impact of masking on the FreeMedQA dataset by progressively removing increasing percentages of the prompt. At 100% masking, all models' free-response performance plummeted to 0.15%. In contrast, their multiple-choice performance remained well above the probability of random chance, hinting at an innate process at work to allow the models to score highly on multiple-choice datasets.